\pdfoutput=1

\ifdefined\REVIEW
    \documentclass[
        a4paper,
    ]{article}
    \usepackage{lineno, setspace}
    \modulolinenumbers[5]
    \pagewiselinenumbers
\else
    \documentclass[
        a4paper,
        twocolumn
    ]{article}
\fi

\usepackage{IEK10} \usepackage{natbib}

\usepackage{todonotes} \makeatletter
\renewcommand{\todo}[2][]{
  \tikzexternaldisable\@todo[#1]{#2}\tikzexternalenable
}
\makeatother

\def\SVT{
  RWTH Aachen University,
  Process Systems Engineering (AVT.SVT),
  Aachen 52074,
  Germany
}

\newcommand{\mytitle}{
    Least Squares and Marginal Log-Likelihood Model Predictive Control using Normalizing Flows
}

\newcommand{\affil}{
  \begin{itemize}[leftmargin=3mm, itemsep=0mm]
    \item[$^a$]\SVT
\end{itemize}
}

\def\firstAuthor{Eike Cramer}
\newcommand{\myauthor}{
\firstAuthor$^{a,*}$\orcidlink{0000-0002-6882-5469}
}
\newcommand{\correspondingauthor}{E. Cramer, \SVT \newline E-mail: eike.cramer@alumni.tu-berlin.de}
\author{\myauthor}

\usepackage[
  colorlinks,
  linkcolor=blue,
  citecolor=blue,
  urlcolor=blue,
  pdftitle={\mytitle},
  pdfauthor={\firstAuthor}
]{hyperref}
\usepackage[capitalise, nameinlink]{cleveref}
\crefname{table}{Tab.}{Tab.}

\usepackage{orcidlink}

\newcommand{\setpgfexternalcounter}[1]{
  \makeatletter \pgfkeysgetvalue{/tikz/external/figure name}\myexternalname
  \expandafter\gdef\csname c@tikzext@no@\myexternalname\endcsname{#1}\makeatother
}

\begin{document}

\ifx\REVIEW\undefined
\twocolumn[
\begin{@twocolumnfalse}
\fi
  \thispagestyle{firststyle}

  \begin{center}
    \begin{large}
      \textbf{\mytitle}
    \end{large} \\
    \myauthor
  \end{center}

  \vspace{0.5cm}

  \begin{footnotesize}
    \affil
  \end{footnotesize}

  \vspace{0.5cm}

  \textbf{Abstract: }
Real-world (bio)chemical processes often exhibit stochastic dynamics with non-trivial correlations and state-dependent fluctuations. 
Model predictive control (MPC) often must consider these fluctuations to achieve reliable performance. 
However, most process models simply add stationary noise terms to a deterministic prediction. 
This work proposes using conditional normalizing flows as discrete-time models to learn stochastic dynamics.
Normalizing flows learn the probability density function (PDF) of the states explicitly, given prior states and control inputs.
In addition to standard least squares (LSQ) objectives, this work derives a marginal log-likelihood (MLL) objective based on the explicit PDF and Markov chain simulations.
In a reactor study, the normalizing flow MPC reduces the setpoint error in open and closed-loop cases to half that of a nominal controller. 
Furthermore, the chance constraints lead to fewer constraint violations than the nominal controller. 
The MLL objective yields slightly more stable results than the LSQ, particularly for small scenario sets.

\vspace*{0.5cm}

  \vspace*{5mm}
\ifx\REVIEW\undefined
\end{@twocolumnfalse}
]
\fi

\section{Introduction}
Model predictive control (MPC) has been established as a key technology in modern control theory and industrial application~\citep{rawlings2017model}.
As the name suggests, MPC requires process models to predict and optimize the process response to control inputs. 
While mechanistic process models are often derived as ordinary differential equations (ODEs), i.e., in continuous time format~\citep{graham2022modelingbook}, MPC is often formulated for using discrete-time state space models~\citep{rawlings2017model}. 
In their standard form, discete-time models predict the system states at the time step $\mathbf{x}[k+1]$ as a function $f$ of the states $\mathbf{x}[k]$, the control inputs $\mathbf{u}[k]$ at the $k$-th time step, and system disturbance $\bm{\omega}$: 
\begin{equation}\label{Eq: DTM definition introduction}
    \begin{aligned}
        \mathbf{x}[k+1] &= f(\mathbf{x}[k], \mathbf{u}[k], \bm{\omega}) \\ 
        \mathbf{x}[0]   &= \mathbf{x}_0
    \end{aligned}
\end{equation}
Here, $\mathbf{x}_0$ is the initial state, and $k\in \mathbb{N}$ is a discrete number indicating the sample time $t=k\Delta t$, where $\Delta t$ is the discrete time step. 
Such models either result from the discretization of continuous-time models or are estimated from process data.
Thus, classic statistical models are some of the earliest adoptions of machine learning in chemical engineering~\citep{Sderstrm2002}. 
Recently, machine learning models such as artificial neural networks have become popular choices of state space models with extensions to Lipschitz ANNs~\citep{Tan2024LipschitzANN}, physics-informed neural networks \citep{Zheng2023PINN_MPC}, and hybrid models \citep{Wu2020hybrid}.
For a review of neural networks-based process models for MPC, the reader is referred to \cite{Ren2022reviewANNcontrol}.

Real-world process data often exhibits a stochastic behavior, e.g., from measurement noise or disturbances. 
Furthermore, some processes are stochastic in nature, i.e., they exhibit inherent stochastic dynamics~\citep{Mesbah2022ml4MPCunderUncertainty}.
This stochastic behavior requires specialized modeling tools that describe the stochastic dynamics of the process~\citep{Varshney2022stateestimationNONGauss}.
Still, most works using state space models rely on deterministic modeling, i.e., the fitted models predict only the most likely realization of the system's states.
Meanwhile, stochastic process models can learn and predict the dynamics of inherently stochastic processes and quantify the impact of measurement noise on the prediction. 
Using stochastic process models, nominal MPC can be extended to stochastic MPC, which formulates a stochastic program~\citep{birge2011introduction} to minimize the expected mean and find optimal controls under uncertainty~\citep{mesbah2016smpc_perspective, rawlings2017model}. 
In a scenario-based stochastic MPC, the stochastic process model generates scenarios of possible realizations to solve the stochastic program via sample average approximation (SAA)~\citep{Shapiro2009SAA}. 
Similarly, stochastic MPC can also be formulated using chance constraints that ensure feasible control actions with a set probability \citep{Schwarm1999chanceconstrainedMPC, Paulson2017MPC_joint_chance_constraints}. 
Note that control theory also knows robust MPC with guarantees for constraint satisfaction~\citep{Morari2007survey_robustMPC}, but the scope of this work is restricted to the stochastic and probabilistic cases.

Most established stochastic process models use an additive, unmeasured noise term, i.e., they describe the stochastic behavior by adding a state-independent zero-mean noise term to a deterministic prediction~\citep{box1967models, mesbah2016smpc_perspective, rawlings2017model}.
The additive noise assumption is independent of the states, excluding any correlation between the states and the stochastic components of the dynamic response. 
For fermentation processes, for instance, the noise intensity varies with the cell concentration~\citep{alvarez2018sdeCSTR}. 
For non-Gaussian distributions, moment-matching techniques have shown promising results, either in single dimensions~\citep{Calfa2014momentmatching_scenario_tree} or in multivariate combinations using Copulas~\citep{Bounitsis2022momentmatching_copula_scenarios}. 
These moment-matching techniques assume that the distribution is fully described by its first four statistical moments.
However, stochastic MPC formulations are multi-stage decision problems requiring multistage scenarios~\citep{Ruszczyski2009multistage_decision_problems}, which are not naturally supported by the copula-moment-matching approach. 
In summary, there is a need for stochastic process models that can describe complex system behaviors, including nonlinear dynamics, potentially non-Gaussian distributions, and state-dependent stochastic dynamic response.

This work uses the deep generative model called normalizing flows \citep{papamakarios2021normalizing} as a probabilistic discrete-time model in the state space. 
Normalizing flows are flexible probability distribution models for multivariate random variables that learn explicit expressions of the probability density function (PDF) using invertible neural networks (INNs) \citep{papamakarios2021normalizing}. 
The basic concept of normalizing flows can be extended to learn conditional PDFs by augmenting the INN with external inputs~\citep{winkler2019learning, rasul2021multivariate, cramer2022normalizing}. 
Thus, normalizing flows also function as multivariate regression models for random variables, i.e., normalizing flows can be used as discrete-time state-space models for multivariate dynamics of systems with correlated, non-Gaussian, and state-dependent noise. 
Notably, normalizing flows learn high-dimensional conditional PDFs without prior assumptions and, thus, provide a highly flexible modeling architecture for stochastic processes. Hence, the control engineer does not need to have detailed knowledge about the system, its dynamics, or its stochastic behavior, as the normalizing flow can, in theory, learn any nonlinear dynamic response in combination with any probability distribution. 
The author has previously used normalizing flows for scenario generation of power generation time series \citep{cramer2022principal, cramer2022normalizing} and electricity prices \citep{cramer2023multivariate, cramer2024dayahead_NF}. 
\cite{Kevrekidis2022normalizingflows_temporalevolution} present a first study investigating the temporal evolution of multivariate densities, i.e., a simulation of stochastic processes using normalizing flow similar to the method proposed in this work. 
Still, the most common use case for normalizing flows is image generation \citep{dinh2015nice, dinh2017density, grathwohl2018ffjord}. 

Normalizing flows sample scenarios of process realizations that can be used to solve stochastic MPC problems. 
Furthermore, this work shows that the explicit expression for the conditional PDF of the system states can be used to formulate a likelihood-based setpoint-tracking objective, i.e., the optimizer maximizes the likelihood that the control inputs achieve the desired setpoints. 
Besides the setpoint-tracking objectives, normalizing flows naturally support formulating chance constraints for any inequality constraints on the process. 
In summary, the contribution of this work is twofold. 
First, this work introduces normalizing flows as state space models for chemical processes with a particular focus on stochastic dynamics.  
Second, this work investigates the likelihood-based MPC objective compared to the established least squares formulation. 
To limit the scope of this paper, the author opts to constrain the analysis to cases where data is available in abundance or mechanistic simulations of the systems are available.

The remainder of this work is organized as follows: 
Section~\ref{sec: normalizing flows} reviews the basic concept of normalizing flows and their extension to conditional probability distributions. 
Section~\ref{sec: probabilistic regression} derives the probabilistic state space formulation using the normalizing flow. 
Next, Section~\ref{sec: Model predictive control using normalizing flows} reviews the least squares MPC formulation and introduces the probabilistic MPC formulation using the explicit PDF expression learned using the normalizing flow. Furthermore, the section shows how to formulate chance constraints from normalizing flow-based system simulations. 
Section~\ref{sec: LV evaluation} analyses the performance of the normalizing flow to predict the stochastic dynamics of an autonomous Lotka-Volterra type system.
Section~\ref{sec: CSTR eval} applies the normalizing flow to learn the stochastic dynamics of a continuous stirred tank reactor (CSTR) and runs MPC in both open- and closed-loop settings. 
Finally, Section~\ref{sec: conclusion} concludes this work.

\section{Normalizing Flows}\label{sec: normalizing flows}
This section introduces the general concept of probability density estimation using normalizing flows and reviews the extension of normalizing flows to conditional probability distributions.

\subsection{Density Estimation using Normalizing Flows}\label{ssec: NF intro}
Normalizing flows are explicit multivariate probability distribution models, i.e., they provide an explicit expression for the PDF $p_X(\mathbf{x})$ of a multivariate random variable $X$ \citep{papamakarios2021normalizing}. 
The random variable $X$ is described as a diffeomorphism $T(\mathbf{z})$ of a standard Gaussian $Z\sim\phi(\mathbf{z}) = \mathcal{N}(\bm{0}, \bm{I})$, i.e., a bijective transformation where both forward and inverse are differentiable at least once:
\begin{equation}
    \begin{aligned}
        T: \; & \mathbb{R}^{D}  \rightarrow \mathbb{R}^{D} \\
           & \mathbf{z} \mapsto T(\mathbf{z}) 
    \end{aligned}
\end{equation}
Here, $\phi$ is the standard Gaussian PDF, and $Z$ must have the same dimensionality $D$ as $X$.  
The diffeomorphism $T$ describes a change of variables. Hence, the PDF of the random variable $X$ is given by the change of variables formula \citep{papamakarios2021normalizing}:
\begin{equation}\label{Eq: change of variables general}
    p_{X}(\mathbf{x} ) = \phi\left(T^{-1}(\mathbf{x})\right) \left| \det \mathbf{J}_{T^{-1}}(\mathbf{x}) \right|
\end{equation}
In Equation~\eqref{Eq: change of variables general}, $\mathbf{J}_{T^{-1}}$ is the Jacobian matrix of the inverse of $T$:
\begin{equation}
    \begin{aligned}
        T^{-1}: \; & \mathbb{R}^{D}  \rightarrow \mathbb{R}^{D} \\
           & \mathbf{x} \mapsto T^{-1}(\mathbf{x}) 
    \end{aligned}
\end{equation}
This inverse transformation projects the random variable $X$ to a space where its counterpart $Z$ is Gaussian.
Hence, Equation~\eqref{Eq: change of variables general} describes the likelihood that a data point $\mathbf{x}$ is part of the probability distribution $p_{X}$.
In practice, the diffeomorphism is parameterized as an INN $\mathbf{x} = T_\Theta(\mathbf{z})$ with inverse $\mathbf{z} = T_\Theta^{-1}(\mathbf{x})$ and parameters $\Theta$~\citep{dinh2017density}. 
The log of Equation~\eqref{Eq: change of variables general} can be used to formulate an expected log-likelihood loss function to train the parameters of the INN \citep{papamakarios2021normalizing}:
\begin{equation}
    \label{Eq: Expected Log-loss}
    \underset{\Theta}{\max}~ \mathbb{E}_X \left[\log~p_{X}(\mathbf{x} ; \Theta)\right]
\end{equation}
The training then maximizes the likelihood that the inverse INN $T_\Theta^{-1}$ maps the random variable $X$ to a standard Gaussian. 
To fit a normalizing flow to a dataset, i.e., a set of discrete samples $\mathbf{x}_i$ of $X$, the expectation in Equation~\eqref{Eq: Expected Log-loss} is substituted with its sample mean:
\begin{equation}
    \label{Eq: Sample Mean Log-loss}
    \underset{\Theta}{\max}~ \frac{1}{N}\sum_{i=1}^N  \log~p_{X}(\mathbf{x}_i ; \Theta)
\end{equation}
Here, N is the number of data points. 
In the following, the subscript $\Theta$ is omitted for the sake of clarity. 
Normalizing flows can learn the probability distribution of any multivariate continuous random variable, including multi-modal or heavily skewed probability distributions~\citep{papamakarios2021normalizing, dinh2017density}. Previous work by the author has also reported good performance on heavily tailed distributions~\citep{cramer2023multivariate}.
The training via likelihood maximization is statistically consistent and asymptotically efficient~\citep{rossi2018mathematical}. 
Hence, normalizing flows converge to the correct PDF for an infinite sample size.

During training, the Jacobian determinant in Equation~\eqref{Eq: change of variables general} has to be computed in every iteration, making the training prohibitively expensive for non-specialized INN architectures \citep{dinh2015nice, dinh2017density}. 
The most popular implementation of INNs is the real non-volume preserving transformation (RealNVP) \citep{dinh2017density} that builds powerful transformations using compositions of partial transformations. These partial transformations are designed to yield triangular Jacobians for efficient determinant computation via the product of the diagonal entries. 
The author elects to omit a detailed discussion of RealNVP to avoid redundancy with existing literature.
For details on the RealNVP architecture and implementation, the reader is referred to the original paper by Dinh~et~al.~\citep{dinh2017density}, the review by Papamakarios~et~al.~\citep{papamakarios2021normalizing}, and the author's previous works on normalizing flows \citep{cramer2022principal, cramer2022normalizing, cramer2023multivariate, cramer2024dayahead_NF}.

\subsection{Conditional Normalizing Flows}\label{ssec: conditional NF}
The standard form of normalizing flows discussed in Section~\ref{ssec: NF intro} can be augmented by including external features. 
Thus, normalizing flows can also describe conditional PDFs $p_{X\vert Y}(\mathbf{x}\vert\mathbf{y})$ depending on an external feature variable $Y$ \citep{winkler2019learning, cramer2022normalizing}. 
For conditional distributions, the INN is augmented with the conditional information as additional inputs \citep{rasul2021multivariate, cramer2022normalizing}:
\begin{equation}
    \begin{aligned}
        T: \; & \mathbb{R}^{D}\times \mathbb{R}^{M}  \rightarrow \mathbb{R}^{D} \\
           & \mathbf{z}, \mathbf{y} \mapsto T(\mathbf{z}, \mathbf{y}) 
    \end{aligned}
\end{equation}
Here, the dimensionality of the conditional variable $Y$ $M$ may be different from the data dimension $D$.

\begin{figure*}
    \centering
    \includegraphics[width=0.7\textwidth]{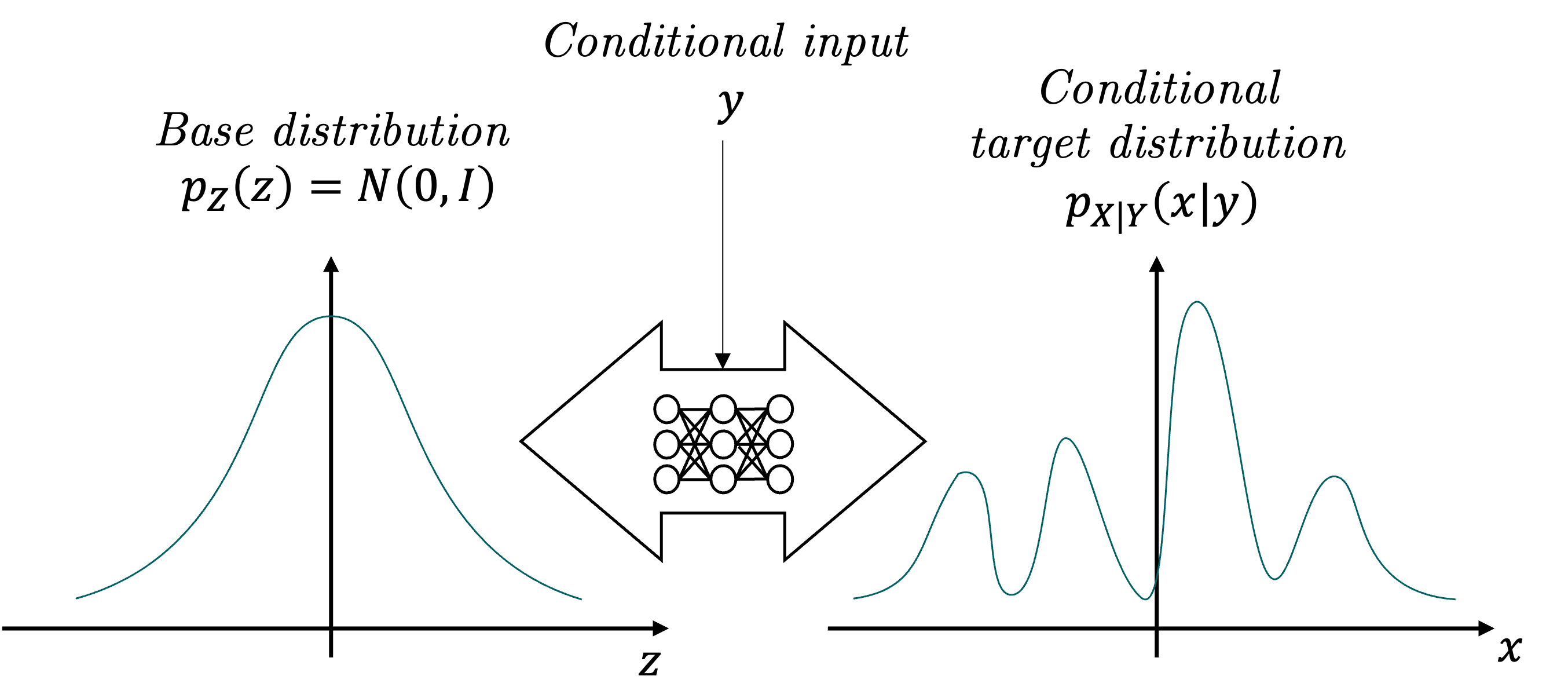}
    \caption{Normalizing flow transformation between random variable $X$ and Gaussian $Z$ with conditional input variable $Y$. The conditional information is added to the INN as proposed in \citep{cramer2022normalizing}. 
    The figure is similar to \cite{cramer2024dayahead_NF}.}
    \label{fig: conditional NF}
\end{figure*}
The conditional variable is considered as an additional input to the INN for both the forward and the inverse transformation:
\begin{align*}
    \mathbf{x} &= T(\mathbf{z}, \mathbf{y}) \\ 
    \mathbf{z} &= T^{-1}(\mathbf{x}, \mathbf{y}) 
\end{align*}
Again, $\mathbf{x}$ are samples from the probability distribution the normalizing flow aims to learn, i.e., the data, $\mathbf{z}$ are samples of the multivariate standard Gaussian, and $\mathbf{y}$ is the numerical conditional information. 

The change of variables described by the conditional normalizing flow remains with respect to the dimensions of the data only. 
The conditional information simply shapes the transformation. 
Thus, the change of variables reads similarly to Equation~\eqref{Eq: change of variables general}:
\begin{equation}\label{Eq: change of variables conditional}
    p_{X\vert Y}(\mathbf{x} \vert \mathbf{y}) = \phi\left(T^{-1}(\mathbf{x}, \mathbf{y})\right) \left| \det \mathbf{J}_{T^{-1}}(\mathbf{x}, \mathbf{y}) \right|
\end{equation}
Again, $\mathbf{J}_{T^{-1}}(\mathbf{x}, \mathbf{y})$ is the Jacobian with respect to $\mathbf{x}$.

After training, samples of the Gaussian can be transformed using the forward transformation to generate new data or scenarios.
\begin{equation}
    \mathbf{\hat{x}}_{i,s} = T(\mathbf{\hat{z}}_s, \mathbf{\hat{y}}_i)
\end{equation}
Here, the $\hat{}$ indicates new samples that are not present in the training or test sets. 
In the following, all normalizing flows are implemented using RealNVP with the adaptation to conditional random variables presented in \cite{cramer2022normalizing}.

\section{Probabilistic State Space Regression}
\label{sec: probabilistic regression}
This section formulates a probabilistic regression model based on the conditional normalizing flow.

Similar to Equation~\eqref{Eq: DTM definition introduction}, the proposed model is written in the Markovian case, i.e., the functional relation only depends on the current time step $\mathbf{x}[k]$ and the control inputs $\mathbf{u}[k]$ at the given time step $k$. 
However, the formulation used in this work predicts the difference between states, i.e., the state increment $\mathbf{x}^{+}[k]=\mathbf{x}[k+1]-\mathbf{x}[k]$:
\begin{equation}\label{eq: generic difference inclusion description}
    \mathbf{x}^+[k] \in \mathbf{F}(\mathbf{x}[k], \mathbf{u}[k])
\end{equation}
This work opts for the state increment vector $\mathbf{x}^{+}[k]$ over the full states $\mathbf{x}[k+1]$ to design a simpler learning problem. 
Equation~\eqref{eq: generic difference inclusion description} presents the difference inclusion description \citep{rawlings2017model} of the probabilistic regression problem, where $\mathbf{F}$ is a set-valued function representing the sample space of the predicted distribution.
The set described by the set-valued function $\mathbf{F}$ can be written as follows:
\begin{equation}
    \mathbf{F} = \left\{f(\mathbf{x}[k], \mathbf{u}[k], \bm{\omega}) \vert  \bm{\omega} \in \bm{\Omega}(\mathbf{x}[k], \mathbf{u}[k])\right\}
\end{equation}
Opposed to typical assumptions in regression and state space models \citep{Sderstrm2002, rawlings2017model}, this work allows for state and control dependencies of the disturbances $\bm{\Omega}(\mathbf{x}[k])$ as well as non-Gaussian distributions.

In the following, Section~\ref{ssec: probabilistic regression using nf} introduces the probabilistic state space regression using conditional normalizing flows. 
Section~\ref{ssec: Multistage Sampling} explains the scenario generation as well as the SAA of integrals.

\subsection{Probabilistic Regression using Normalizing Flows}
\label{ssec: probabilistic regression using nf}
This work uses the conditional normalizing flow to describe the probability distribution over the set-valued function in Equation~\eqref{eq: generic difference inclusion description}. 
For the sake of brevity, the time indication $[k]$ is omitted for the remainder of this subsection. 
\begin{figure*}
    \centering
    \includegraphics[width=0.7\textwidth]{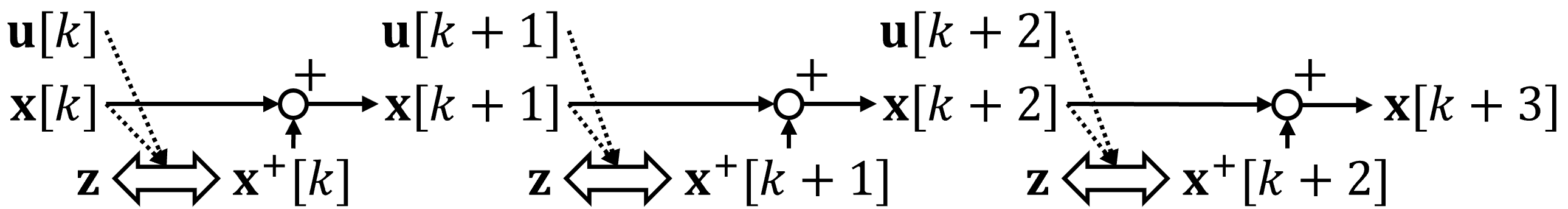}
    \caption{
    Three steps of regression using normalizing flows. 
    Here, $\mathbf{x}[k]$ are the states, $\mathbf{u}[k]$ are the control inputs, and $\mathbf{z}\sim \mathcal{N}(\mathbf{0}, \mathbb{I})$ is a multivariate Gaussian with $\dim(\mathbf{x}) = \dim(\mathbf{z})$. 
    The two-headed arrow represents the INN $T_R$ (Equation~\eqref{Eq: INN regression}) with conditional inputs mapping between $\mathbf{z}$ and the increment distribution $\mathbf{x}^+[k]$. 
    }
    \label{fig: NF regression scheme}
\end{figure*}

The principle of training normalizing flows described in Section~\ref{ssec: NF intro} and \ref{ssec: conditional NF} translates to any combination of continuous random variables $X$ and conditional inputs $Y$. Thus, normalizing flows can be used as probabilistic, multivariate regression models.
For the regression in state space, the transformation is constructed to be diffeomorphic between the target variable, i.e., the state increment $\mathbf{x}^+$, and the Gaussian variable $\mathbf{z}$.
The combination of the current states $\mathbf{x}$ and the controls $\mathbf{u}$ form the conditional information $\mathbf{y}$ in Equation~\eqref{Eq: change of variables conditional}:
\begin{equation}\label{Eq: INN regression}
    \mathbf{x}^+ = T_R \left(\mathbf{z},\underbrace{\left[\mathbf{x}, \mathbf{u}\right]}_{=\mathbf{y}}\right)
\end{equation}
Here, $T_R$ is the INN with subscript $R$ for regression. In practice, the state vector $\mathbf{x}$ and the controls $\mathbf{u}$ are concatenated to form a joined conditional input vector $\mathbf{y}$.
Figure~\ref{fig: NF regression scheme} visualizes the normalizing flow-based regression scheme for three steps. The two-headed arrows indicate the INN $T_R$.

Using the difference inclusion formulation \citep{rawlings2017model}, the normalizing flow-based regression model may be written analog to Equation~\eqref{eq: generic difference inclusion description}:
\begin{equation}
    \mathbf{x}^+ \in \mathbf{T_R} \left(\mathbf{x}, \mathbf{u}\right)
\end{equation}
Here, $\mathbf{T_R}$ is the set of possible realizations described by the normalizing flow.
The inverse of \eqref{Eq: INN regression} reads:
\begin{equation}\label{Eq: INN regression inverse}
    \mathbf{z} = T_R^{-1} \left(\mathbf{x}^+,\left[\mathbf{x}, \mathbf{u}\right]\right)
\end{equation}
Note that RealNVP \citep{dinh2017density} requires the number of states to be $\geq 2$ but poses no limitations on $\mathbf{u}$, i.e., the inputs $\mathbf{u}$ may be continuous or discrete. 

The PDF of the predicted time step is again explicitly described via the change of variables formula:
\begin{equation}\label{Eq: change of variables regression}
\begin{aligned}
    p(\mathbf{x}^+& \vert \left[ \mathbf{x}, \mathbf{u} \right]) =   \\&\phi\left(T_R^{-1}\left(\mathbf{x}^+,\left[\mathbf{x}, \mathbf{u}\right]\right)
    \right) 
    \left|\det \mathbf{J}_{T_R^{-1}} \left(\mathbf{x}^+,\left[\mathbf{x}, \mathbf{u}\right]\right) 
    \right|
\end{aligned}
\end{equation}

\subsection{Multistep Sampling and Integral Approximation}\label{ssec: Multistage Sampling}
Equation~\eqref{Eq: INN regression} describes the general regression via the conditional normalizing flow. 
As the variable $\mathbf{z}$ is a random variable, the expression cannot be evaluated for the full distribution of $\mathbf{z}$. 
However, such evaluations are necessary to simulate the system for multiple steps or to compute integrals, e.g., over an objective. 
The system is written in discrete time, and the expectation over the states can be approximated via SAA over $N_S$ scenarios $\mathbf{x}_s[k]$. 
Thus, the integral over a function $J(\mathbf{x}(t))$ can be approximated as follows~\citep{Asmussen2007}:
\begin{equation}\label{eq: MCMC function integral}
    \int_0^T\mathbb{E}_\mathbf{x}[J(\mathbf{x}(t))]\ dt \approx 
    \frac{\Delta t}{N_S}\sum_{k=1}^{N_T}\sum_{s=1}^{N_S} J(\mathbf{x}_s[k]) 
\end{equation}
Here, $\Delta t$ is the discrete time step, and $N_T$ is the total number of steps.

To obtain the scenarios to compute Equation~\eqref{eq: MCMC function integral}, this work uses a Markov Chain Monte Carlo (MCMC)~\citep{Asmussen2007} to compute approximations of the distribution over the states and time. 
Starting at an initial state, MCMC initiates $N_S$ Markov Chains that are simulated for $N_T$ steps. 
The single step prediction for scenario $s$ in the $k$-th step then reads:
\begin{equation}\label{eq: MCMC step}
    \mathbf{x}^{+}_{s}[k] = T_R (\mathbf{z}_{s,k}, \left[\mathbf{x}_{s}[k], \mathbf{u}[k]\right])
\end{equation}
In Equation~\eqref{eq: MCMC step}, $\mathbf{x}^{+}_{s}[k]$ is the scenario $s$ state increment at the $k$-time step, and $\mathbf{z}_{s,k}$ is a sample from the multivariate normal base distribution for scenario $s$ and time step $k$. 
Iterating over Equation~\eqref{eq: MCMC step} yields $N_S$ trajectories of possible scenarios that can be used to evaluate integrals as in Equation~\eqref{eq: MCMC function integral}.

\section{Model Predictive Control using Normalizing Flows}
\label{sec: Model predictive control using normalizing flows}

This section discusses least squares and likelihood-based state-tracking objectives for MPC based on the normalizing flow state space model. 
Notably, such MPC formulations with random variables present multi-stage stochastic programs, which can be difficult to formulate and solve~\citep{Shapiro2009SAA}.
Thus, the analysis in this paper is restricted to single-stage approximations of the multi-stage stochastic programs to limit the range of this work.
Hence, there are no scenario trees or recourse decisions, and all MPC formulations in this work compute a single set of control inputs $\{\mathbf{u}[k]\}_{\forall k=1,\dots, N}$ for the considered time horizon.

The content of the section is as follows:
First, Section~\ref{ssec: MSE obj} reviews the established formulation using the least squares objective. 
Second, Section~\ref{ssec: likelihood obj} introduces an objective based on the conditional log-likelihood described by the normalizing flow.
Finally, Section~\ref{ssec: chance constraints via quantiles} shows how chance constraints can be formulated using the MCMC scenarios. 

\subsection{Least Squares Objective}\label{ssec: MSE obj}
Setpoint MPC computes the controls $\mathbf{u}$ by minimizing the predicted offset between the system setpoints $\mathbf{x}^*[k]$ and the states $\mathbf{x}[k]$ over the given time horizon $k=1,\dots, N_T$, where $N_T$ is the number of time steps.
The most common formulation for the objective function is the least squares (LSQ) formulation~\citep{rawlings2017model}.
A stochastic model yields a distribution of system realizations and, thus, stochastic MPC minimizes the expected LSQ between the states and their setpoints \citep{mesbah2016smpc_perspective, rawlings2017model}. 
The LSQ MPC problem using the normalizing flow $\mathbf{T_R}$ then reads:
\begin{equation}\label{Eq: stochastic MPC objective}
\begin{aligned}
    \underset{\forall\mathbf{u}[k]}{\min} ~& \sum_{k=0}^{N_T-1}\mathbb{E}\left[\Delta\mathbf{x}[k]^T\mathbf{Q}\Delta\mathbf{x}[k] \right]\Delta t\\
    \text{s.t.}~&
    \begin{aligned}
    \mathbf{x}^+[k] &\in \mathbf{T_R} \left(\mathbf{x}[k], \mathbf{u}[k]\right) \\
    \Delta \mathbf{x}[k] &= \mathbf{x}^*[k]-\mathbf{x}[k] \\ 
    \mathbf{x}[k+1] &= \mathbf{x}[k] + \mathbf{x}^+[k]
    \end{aligned}
\end{aligned}
\end{equation}
Here, $\Delta\mathbf{x}[k]$ is the setpoint error at time $k$, and $\mathbf{Q}$ is a diagonal weight matrix.
In practice, the LSQ objective in Equation~\eqref{Eq: stochastic MPC objective} is intractable and instead is approximated by the expectation interval given by Equation~\eqref{eq: MCMC function integral}.

The LSQ formulation in Equation~\eqref{Eq: stochastic MPC objective} is written for the case where all states $\mathbf{x}$ have setpoints. 
Naturally, the formulations may be modified for cases where only a subset of the states should follow setpoints by including only those states in the formulation. In this case, the weight matrix $\mathbf{Q}$ includes zero entries on its diagonal.

\subsection{Likelihood Objective}
\label{ssec: likelihood obj}
The normalizing flow-based regression model introduced in Section~\ref{ssec: probabilistic regression using nf} gives an \emph{explicit} expression for the conditional PDF that allows for the computation of the likelihood of the controls inducing the desired state increments. 
Thus, this work proposes a substitute for Equation~\eqref{Eq: stochastic MPC objective} with a log-likelihood maximization. 
The maximum log-likelihood objective reads:
\begin{equation}\label{Eq: max log p MPC}
\begin{aligned}
    \underset{\forall\mathbf{u}[k]}{\max} &~ \sum_{k=0}^{N_T-1}\mathbb{E}\left[\log p\left(\mathbf{x}^{+*}[k]\vert \left[ \mathbf{x}[k], \mathbf{u}[k] \right]\right) \right] \Delta t \\
    \text{s.t.}~&
    \begin{aligned}
        \mathbf{x}^{+*}[k] &= \mathbf{x}^*[k+1]-\mathbf{x}[k] 
    \end{aligned}
\end{aligned}
\end{equation}
Here, $\mathbf{x}^{+*}[k]$ may be viewed as the perfect response achieving the setpoint in the next time step, $\mathbf{x}[k]$ are the current states, $\mathbf{u}[k]$ are the control inputs and degrees of freedom, $\mathbf{x}^*[k+1]$ are the setpoints for the next time steps, and $\log p(\cdot\vert\cdot, \cdot)$ is the conditional log-likelihood function described by the change of variables formula in Equation~\eqref{Eq: change of variables regression}. 
Equation~\eqref{Eq: max log p MPC} includes the expectation operator since the current time step $\mathbf{x}[k]$ is a random variable for $k>0$. 
Solving Equation~\eqref{Eq: max log p MPC} then maximizes the likelihood that the system takes the perfect step to the setpoints given the probability distribution described by the normalizing flow and given the respective previous states and the control inputs.

The explicit formulation of the conditional PDF in Equation~\ref{Eq: change of variables regression} gives a single likelihood value for all state increments at a given time $\mathbf{x}^+[k]$.  
To allow for the common case of having setpoints only for a subset of the states, Equation~\eqref{Eq: max log p MPC} must be written for the marginal probability distribution of those states, i.e., via the marginal log-likelihood (MLL):
\begin{equation}\label{eq: MLL MPC}
    \text{MLL} =\mathbb{E}_{\mathbf{x}}\left[ \int_{\mathcal{X}^+_j}\log p\left(
    \begin{bmatrix}
        \mathbf{x}_i^{+*}\\
        \mathbf{x}_j^{+}
    \end{bmatrix}
    \vert \left[ \mathbf{x}, \mathbf{u} \right]\right)\ d\mathbf{x}^+_j \right]
\end{equation}
Here, $ \mathbf{x}_i^{+}$ and $\mathbf{x}_j^{+}$ are the state increments with and without setpoints, respectively, and $\mathcal{X}^+_j$ is the set of $\mathbf{x}_j^{+}$ given the current states $\mathbf{x}$ and controls $\mathbf{u}$. The time dependency $[k]$ is omitted in Equation~\eqref{eq: MLL MPC} for the sake of brevity.  
Both Equation~\eqref{Eq: max log p MPC} and~\eqref{eq: MLL MPC} can be approximated via the MCMC integral approximation in Equation~\eqref{eq: MCMC function integral} using scenarios generated from the normalizing flow.
In the following, the MCMC solution of Equation~\eqref{eq: MLL MPC} is referred to as the MLL objective.

\subsection{Chance Constraints}\label{ssec: chance constraints via quantiles}
Normalizing flows model the joint probability distribution of all dynamic states $\mathbf{x}[k]$ at a given time $k$. 
The knowledge of the probability distribution allows the user to formulate inequality constraints $h(\mathbf{x}[k]) \geq 0$ as chance constraints:
\begin{equation}\label{eq: general CC}
    \text{Pr}\left\lbrace h(\mathbf{x}[k]) \geq 0 \right\rbrace \geq \varepsilon 
\end{equation}
Here, $h(\cdot)$ is the inner constraint, and $\varepsilon$ is the probability value that should hold for satisfying the constraint. 
The function $h$ might be nonlinear, and there exists no closed-form expression to implement Equation~\eqref{eq: general CC} into a numeric program. 
Instead, the chance constraint in Equation~\eqref{eq: general CC} can be replaced with a constraint on the corresponding quantile of the distribution of the function $h$: 
\begin{equation}
    p_{h(\mathbf{x}[k])}(1-\varepsilon) \geq 0 
\end{equation}
In practice, the quantile $p_{h(\mathbf{x}[k])}(1-\varepsilon)$ is computed as an empirical quantile from evaluating $h$ for every scenario generated using MCMC. 
From these scenarios, the quantile is computed via linear interpolation of the empirical inverse cumulative distribution function (CDF)~\citep{Hyndman1996quantilesStatisticalPackages}.
Note that the discrete-time formulation does not guarantee satisfying path constraints.

\section{Simulation of Stochastic Autonomous Systems}\label{sec: LV evaluation}
This section considers the simulation of an autonomous system, namely a stochastic version of the Lotka-Volterra system~\citep{brauer2012mathematical}.
The continuous time system is modeled as an SDE:
\begin{equation}\label{eq: lv conti}
    \begin{aligned}
        dX_t &= f_1 (X_t, Y_t)dt + g_1(X_t, Y_t)dB_t \\
        dY_t &= f_2 (X_t, Y_t)dt + g_2(X_t, Y_t)dB_t \\
        f_1 (X_t, Y_t) &= X_t\left(\alpha-\beta Y_t\right) - \zeta (X_t-\bar{x}) \\
        f_2 (X_t, Y_t) &= X_t\left(\delta X_t-\gamma\right) \\ g_1(X_t, Y_t) & =\sigma X_t^{0.6} Y_t^{0.6} \\
        g_2(X_t, Y_t) & =\sigma Y_t^{1.3} \\
    \end{aligned}
\end{equation}
Here, $B_t$ is Brownian motion. Note that the stochastic components $g_1$ and $g_2$ are nonlinear functions of the states. 
The parameters for the SDE are listed in Table~\ref{tab: LV Parameter}.
The first state drift equation $f_1$ includes the term $- \zeta (X_t-\bar{x})$ which corresponds to a P-control action. 
Without this term, the stochastic Lotka-Volterra system will result in the extinction of both species almost surely. For a detailed discussion of expectations of future system behavior, the reader is referred to \cite{thygesen2023stochastic}.

\begin{table}
\centering
\caption{Parameter values for the Lotka-Volterra system in Equation~\eqref{eq: lv conti}.}
\label{tab: LV Parameter}
\begin{tabularx}{\columnwidth}{@{}lXlX@{}}
\toprule
Parameter & Value            & Parameter & Value \\ \midrule
$\alpha$  & 1.5              & $\zeta$   & 0.01  \\
$\beta$   & 0.01             & $\sigma$  & 0.01  \\
$\gamma$  & 0.3              & $\bar{x}$ & 1000  \\
$\delta$  & 4$\cdot 10^{-4}$ &   \\ \bottomrule
\end{tabularx}
\end{table}

To obtain training data for the normalizing flow, Equation System~\eqref{eq: lv conti} is simulated using the Euler-Murayama method~\citep{thygesen2023stochastic}.
The discrete time interval is set to $\Delta t=0.01$ and the system is simulated for 100 periods. Note that there are multiple periods in a full revolution of the periodic repetition. 
The RealNVP \citep{dinh2017density} normalizing flow is implemented using the python-based machine learning library \emph{TensorFlow} \citep{i2015tensorflow} and trained for 10 epochs using the optimizer Adam \citep{kingma2015adam} and a learning rate of $2\times 10^{-4}$.

\begin{figure}
    \centering
    \includegraphics[width=\columnwidth]{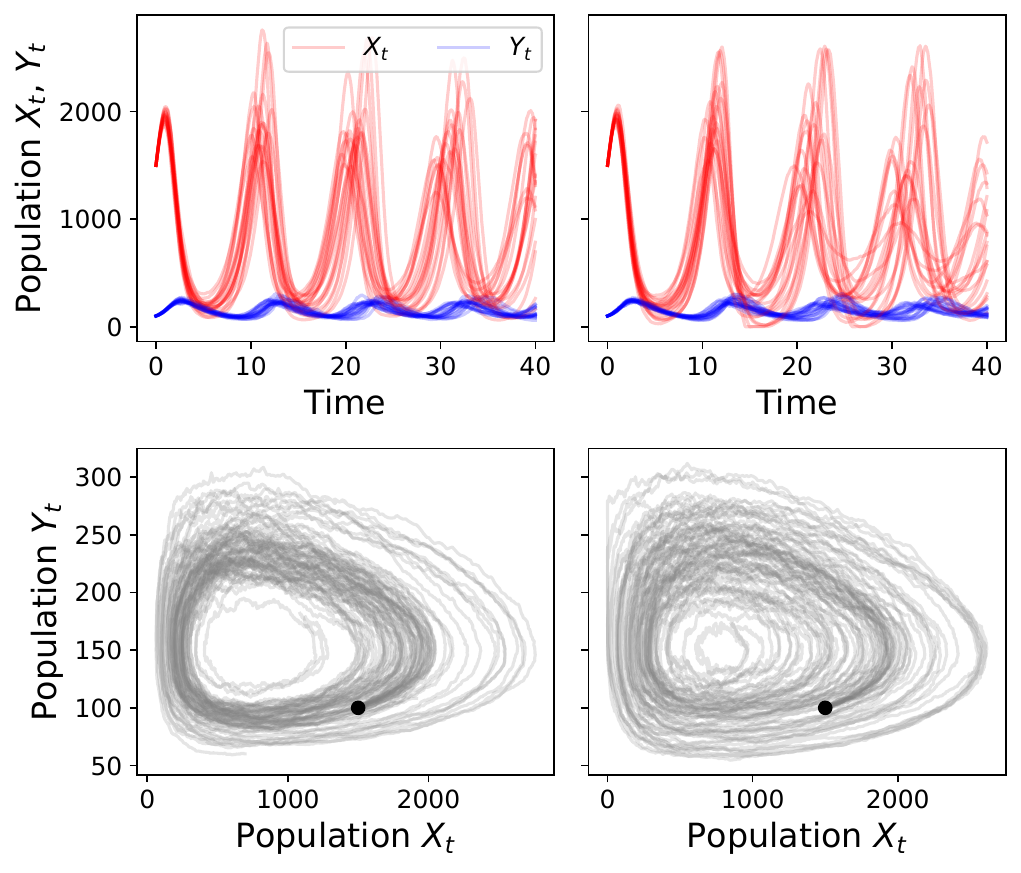}
    \caption{
        Simulation of the Lotka-Volterra system (Equation~\eqref{eq: lv conti}) with the Euler-Murayama simulation (left) and the normalizing flow simulation (right).
        The top row shows the populations' evolutions over time, and the bottom row shows the same in the state space.
        In the bottom row, the initial state is marked with a black dot. }
    \label{fig: LV Sim}
\end{figure}
Figure~\ref{fig: LV Sim} shows the simulation results of the Euler-Murayama method (left) and the simulation using the normalizing flow (right). 
Starting from the initial point $(x_0,y_0)^T=(1500,100)^T$ marked in black in the second row, the figure shows the behavior of 20 scenarios over time (top) and in the state space (bottom). 
Both simulations progress similarly and complete roughly four revolutions of the periodic system. 
The scenarios generated using the normalizing flow appear to reflect the scenarios of the Euler-Murayama solution well. 

\begin{table}
\centering
\caption{
    Mean values $\mu_{ES}$ and standard deviation $\sigma_{ES}$ of energy scores~\citep{gneiting2007strictly} computed for the Lotka-Volterra system over 100 realizations with 200 scenarios each. Values are presented for the Euler-Murayama (EM) and the Normalizing Flow (NF) simulations. 
    }
\label{tab: LV Energy Score}
\begin{tabularx}{\columnwidth}{@{}lXXXX@{}}
\toprule
                   & \multicolumn{2}{c}{$N_T=1000$} & \multicolumn{2}{c}{$N_T=4000$} \\ \midrule
                   & EM             & NF            & EM             & NF            \\
$\mu_{ES}(X_T)$    & 5000           & 5300          & 24000          & 28000         \\
$\sigma_{ES}(X_T)$ & 440            & 450           & 4400           & 4400          \\
$\mu_{ES}(Y_T)$    & 1300           & 1600          & 2100           & 2400          \\
$\sigma_{ES}(Y_T)$ & 80             & 90            & 386            & 380           \\ \bottomrule
\end{tabularx}
\end{table}
Table~\ref{tab: LV Energy Score} shows a quantitative analysis using the energy score~\citep{gneiting2007strictly}. 
The energy score is an evaluation measure for scenario forecasts that is widely used in the forecasting literature. For a definition of the energy score, see Appendix~\ref{app: energy score}.
While, in general, a lower energy score indicated better scenarios, the interest of this evaluation lies in reproducing the behavior of the Euler-Murayama simulation. Thus, the primary concern is whether the energy score values of the normalizing flow are similar to those of the Euler-Murayama simulation, and the absolute values in Table~\ref{tab: LV Energy Score} are less significant. 
The energy scores of the normalizing flow are slightly higher, yet still close to the benchmark, with similar standard deviation over the 100 simulations. 
These results indicate good performance, particularly when considering the long simulation horizons of 1000 and 4000 time steps, respectively. 
In conclusion, the normalizing flow state space model learns and replicates the autonomous Lotka-Volterra system well, which is promising for further analysis of non-autonomous systems and control.

\section{Simulation and Control of a CSTR}\label{sec: CSTR eval}
This section applies the normalizing flow to learn the discrete-time behavior of a CSTR and optimize the controls in open and closed-loop control settings. 
First, Section~\ref{ssec: Cast Study intro} introduces the case study and explains the training setup details for the normalizing flow. 
Next, Section~\ref{ssec: case study simulation} shows a probabilistic CSTR simulation.
Section~\ref{ssec: case study open look mpc} shows the results of an open-loop MPC optimization using the LSQ and the MLL objective.  
Finally, Section~\ref{ssec: CSTR closed loop} draws a comparison in closed-loop performance between the nominal benchmark MPC and the two normalizing flow MPCs.

\subsection{Stochastic CSTR}\label{ssec: Cast Study intro}
\begin{figure}
    \centering
    \includegraphics[width=0.9\columnwidth]{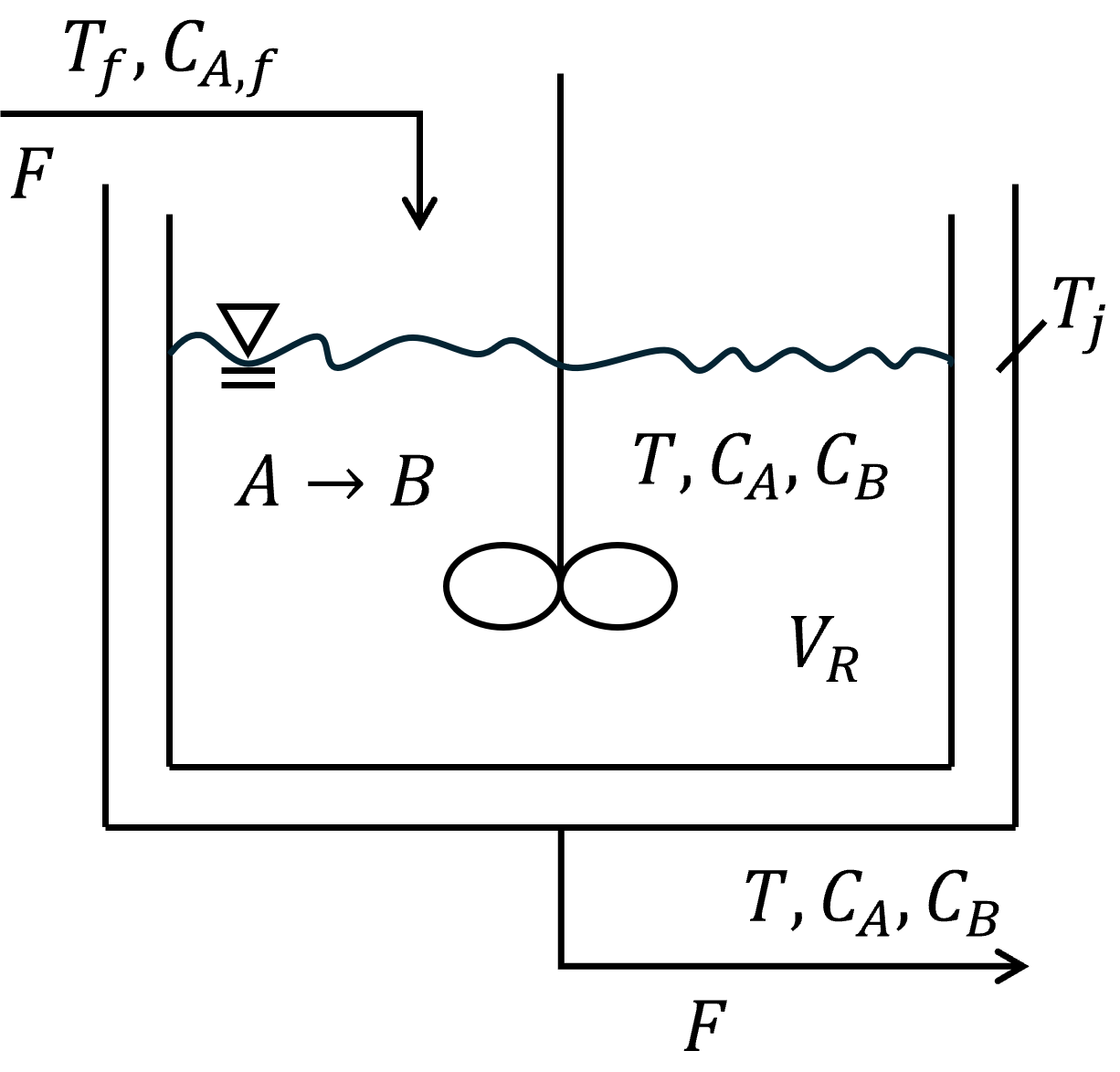}
    \caption{
    Continuously stirred tank reactor (CSTR) sketch similar to \cite{bequette1998process}.
    State variables are the concentration of component A, $C_A$, and the reactor temperature $T$.
    Control variables are the feed temperature $T_f$, the jacket temperature $T_j$, and the feed concentration of component A, $C_{Af}$.
    }
    \label{fig: CSTR Sketch}
\end{figure}

The second case study in this work is the CSTR reactor model by~\cite{bequette1998process}, which presents a non-autonomous system. 
Figure~\ref{fig: CSTR Sketch} shows a sketch of the process. 
The CSTR has two differential states: the reactor temperature $T$ and the concentration $C_A$ of the reagent $A$. 
The two states are controlled via three control inputs: the feed temperature $T_f$, the reactor jacket temperature $T_j$, and the concentration of the reagent $A$ in the feed $C_{Af}$.

\begin{table*}
\centering
\caption{CSTR \citep{bequette1998process} parameters from \cite{MatlabCSTRWebsite}.}
\label{tab: CSTR params}
\begin{tabularx}{\textwidth}{@{}lrlX@{}}
\toprule
Parameter    & Value      & Unit                 & Description                                                 \\ \midrule
$F$          & 1          & m$^3$/h              & Volumetric flow rate                                        \\
$V_R$        & 1          & m$^3$                & Reactor volume                                              \\
$R$          & 1.985      & kcal/(kmol·K)        & Ideal gas constant                              \\
$\Delta   H$ & -5,96      & kcal/kmol            & Heat of reaction per mole                                   \\
$E$          & 11,843     & kcal/kmol            & Activation energy per mole                                  \\
$k_0$        & 34,930,800 & 1/h                  & Pre-exponential nonthermal   factor                         \\
$\rho   C_P$ & 500        & kcal/(m$^3$$\cdot$K) & Density multiplied by heat   capacity                       \\
$UA$         & 150        & kcal/(K$\cdot$h)     & Overall heat transfer coefficient multiplied by tank surface area \\ \bottomrule
\end{tabularx}
\end{table*}

The deterministic component of the dynamics is given by the following ODE: 
\begin{equation}
\label{Eqs: CSTR model}
    \begin{aligned}
        \frac{dC_A}{dt} &= \frac{F}{V_R}\left( C_{Af} - C_A \right) - r \\
        \frac{dT}{dt}   &= \frac{F}{V_R}\left(T_f - T\right) - \frac{\Delta H}{\rho C_P}r - \frac{UA}{\rho C_P V_R}(T-T_j) \\
        r &= k_0 C_A \exp\left(-\frac{E}{RT}\right)
    \end{aligned}
\end{equation}
Here, $C_A$ is the concentration of component A, $T$ is the reactor temperature, $T_f$ is the feed temperature, $T_j$ is the jacket temperature, and $C_{Af}$ is the feed concentration of component A. 
Table~\ref{tab: CSTR params} lists the parameters and their values.
The control variables and their limits are listed in Table~\ref{tab: cstr control limits}.
\begin{table}
\centering
\caption{CSTR control limits \citep{bequette1998process}.}
\label{tab: cstr control limits}
\begin{tabularx}{\columnwidth}{@{}lXX@{}}
\toprule
                      & Min & Max \\ \midrule
$T_j$ [K]             & 273 & 322 \\
$T_f$ [K]             & 273 & 322 \\
$C_{Af}$ [kmol/m$^3$] & 0   & 12  \\ \bottomrule
\end{tabularx}

\end{table}

For the experiments in this work, the CSTR is studied as a stochastic process, i.e., random events impact the dynamics of the process. 
Such stochastic processes are common in bio-based processes such as cell growth or fermentation~\citep{alvarez2018sdeCSTR}. 
To simulate data of the stochastic process, the CSTR is extended to be an SDE, where Equation System~\eqref{Eqs: CSTR model} represents the drift function and the noise term is given by:
\begin{equation}\label{eq: CSTR noise}
    g(C_A, T) = 
    \begin{bmatrix}
        \sigma_{C_A}  C_A \\
        \sigma_{T}  \vert T-270K\vert^{1.5}
    \end{bmatrix}
\end{equation}
Notably, the noise terms have nonlinear dependence on the states, which means that the system cannot be described using standard additive noise terms. 
The parameter values are $\sigma_{C_A}=0.05$ and $\sigma_{T}=0.01$.

The CSTR model is simulated using the Euler-Murayama rule~\citep{thygesen2023stochastic} with a discretization of 5\ minutes per interval to create a dataset to train the normalizing flow. 
For the simulation, control inputs are randomly sampled from a uniform distribution over the control limits listed in Table~\ref{tab: cstr control limits}. 
During the simulation, the control inputs are kept constant for 50\ minutes, i.e., ten steps, for 1000 intervals. 
After simulation, the two states are normalized, and the controls are scaled to $[-1,1]$ with -1 and 1 representing the minimum and maximum control actions in Table~\ref{tab: cstr control limits}, respectively. 
The targets for the normalizing flow regression are created by computing the state increments. 

The normalizing flow uses the same implementation as in Section~\ref{sec: LV evaluation}, is trained for 20 epochs, and has a learning rate of $2\times 10^{-4}$.

\subsection{Simulation of CSTR}
\label{ssec: case study simulation}
In a first analysis, this section evaluates whether the normalizing flow is able to simulate the stochastic state trajectories of the CSTR for a selection of control inputs. 
\begin{figure}
    \centering
    \includegraphics[width=\columnwidth]{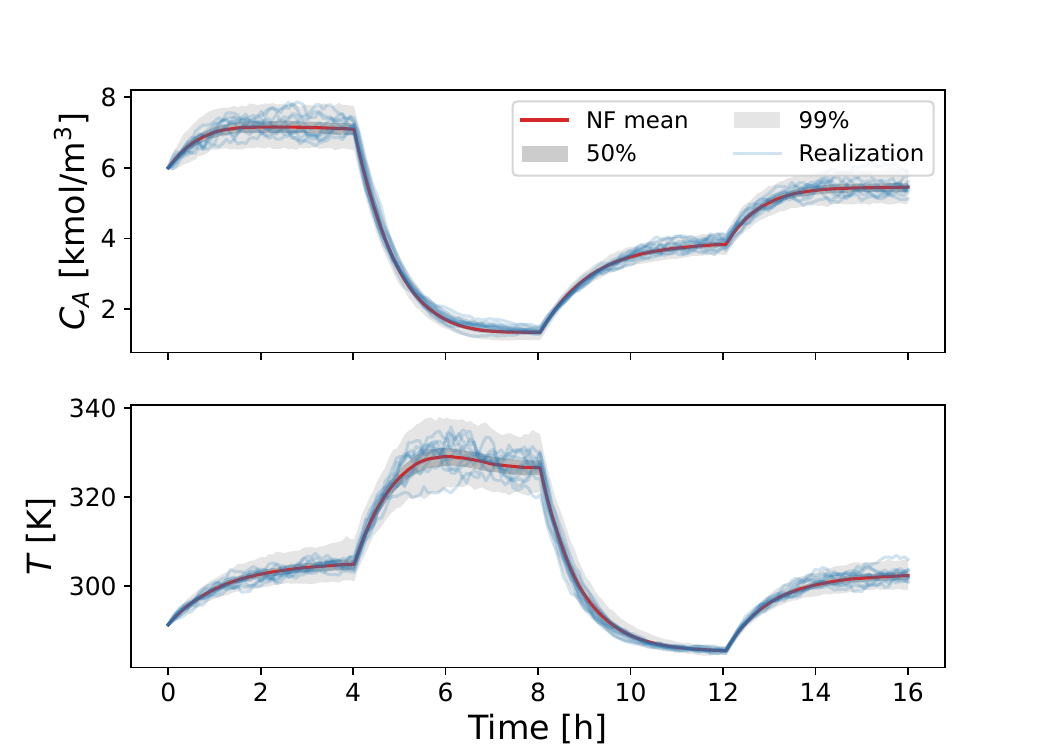}
    \caption{
    Simulation of the reagent concentration $C_A$ and the reactor temperature $T$ for the control inputs in Table~\ref{tab: CSTR Simlation inputs}.
    The first and the second rows show the mean (``NF mean'') and 50\% and 99\% prediction intervals for the normalizing flow in comparison to a simulation using the true model equations in \eqref{Eqs: CSTR model} (``Realization'').
    The prediction intervals are estimated from 1000 scenarios. 
    }
    \label{fig: CSTR NF simulation}
\end{figure}
Figure~\ref{fig: CSTR NF simulation} shows a simulation of the CSTR for 16 1\ h intervals (12\ time steps each) for randomly selected control values. 
The normalizing flow generates 1000 scenarios via MCMC, and the Figure shows the empirical mean values estimated by the normalizing flow (MF mean), the 50\% and 99\% prediction intervals, and ten realization scenarios computed via the Euler-Murayama simulation of Equation System~\eqref{Eqs: CSTR model} and \eqref{eq: CSTR noise}. 
The control inputs are listed in Table~\ref{tab: CSTR Simlation inputs}.
All simulations only receive the initial values and the controls.

\begin{table}
\centering
\caption{Control inputs to the CSTR simulation. Each input is held for four hours, i.e., 48 time steps in the 5-minute discretization. }
\label{tab: CSTR Simlation inputs}
    \begin{tabularx}{\columnwidth}{@{}lXXXX@{}}
    \toprule
    Inputs                & 0-4\,h & 5\,h-8\,h & 9\,h-12\,h & 13\,h-16\,h \\ \midrule
    $T_J$ [K]             & 300   & 330     & 270     & 300     \\
    $T_f$ [K]             & 292   & 292     & 332     & 292     \\
    $C_{Af}$ [kmol/m$^3$] & 8     & 2       & 4       & 6       \\ \bottomrule
    \end{tabularx}
\end{table}

The results in Figure~\ref{fig: CSTR NF simulation} show that the normalizing flow predicts the dynamics of the stochastic CSTR well. 
Almost all realization scenarios are fully enclosed within the 99\% prediction intervals, and the simulation remains stable throughout the 16-hour time interval. The stability of the simulation highlights the normalizing flow's ability to learn the interaction between the noise and the drift of the dynamics of the system. Whenever noise leads to a divergence of the states, the drift is also affected as it, too, depends on the states. 
Notably, prediction intervals for the reactor temperature get wider for times (hours 4 to 8) when the scenarios also show higher variance. After the widening, the prediction intervals become narrower for later times (hours 8 to 12). The ability to reduce the width of the prediction intervals even for long simulation horizons underlines the ability of the normalizing flow to learn both the nonlinear drift of the dynamics and the state dependency of the fluctuations. 

\begin{table}
\centering
\caption{
    Mean values $\mu_{ES}$ of energy scores~\citep{gneiting2007strictly} computed for the non-autonomous CSTR system over 100 realizations with 200 scenarios each for the control inputs presented in Table~\ref{tab: CSTR Simlation inputs}.
    Values are presented for the Euler-Murayama (EM) and the Normalizing Flow (NF) simulations. }
\label{tab: ES CSTR}
\begin{tabularx}{\columnwidth}{@{}lXX@{}}
\toprule
                   & EM   & NF   \\ \midrule
$\mu_{ES}(C_A)$    & 2.0  & 2.2  \\
$\mu_{ES}(T)$      & 21.6 & 21.7 \\ \bottomrule
\end{tabularx}
\end{table}
For a quantitative analysis, Table~\ref{tab: ES CSTR} lists the mean energy score (see Appendix~\ref{app: energy score}) values for the 100 realizations computed for the Euler-Murayana simulations and the normalizing flow. 
As discussed in Section~\ref{sec: LV evaluation}, the aim of the energy score evaluation is to achieve the same values using the normalizing flow as with the Euler-Murayama, which takes the role of the ground truth. 
The energy score values for the simulations are very close for both the concentration $C_A$ and the reactor temperature $T$. 

In summary, the simulation of the stochastic CSTR confirms that the normalizing flow yields stable simulations of stochastic non-autonomous systems with accurate prediction intervals.

\subsection{Open-loop MPC}\label{ssec: case study open look mpc}
This section applies the normalizing flow for open-loop MPC of the stochastic CSTR case study. 
The open-loop MPC solves the optimization of the LSQ formulation in Equation~\eqref{Eq: stochastic MPC objective} and the MLL formulation in Equation~\eqref{Eq: max log p MPC} for a fixed time horizon, respectively. 
In every time step, the stochastic formulation is solved for 100 scenarios. 
The controller aims to achieve four different setpoints after each other. 
The four setpoints are each held for 4\ h, which equals 48 time steps in the discretization to 5\ minutes. The optimizer may freely manipulate the three controls for every time step within the bounds listed in Table~\ref{tab: cstr control limits}. 
The four setpoints are listed in Table~\ref{tab: open-loop control setpoints}.

The open-loop optimization further considers a safety constraint on the reactor temperature that limits the temperature to be below 310\ K. 
The constraint is implemented as a chance constraint with a probability of 90\%:
\begin{equation}\label{eq: CSTR chance constraint}
    Pr\{T\leq 310 K\}\geq 90\%
\end{equation}
The chance constraint is implemented via the quantile replacement discussed in Section~\ref{ssec: chance constraints via quantiles}.

\begin{table}
\centering
\caption{Setpoints for open-loop control. Each setpoint is held for 4\ h, which is 48 control intervals of 5\ minutes each.}
\label{tab: open-loop control setpoints}
\begin{tabularx}{\columnwidth}{@{}lXXXX@{}}
\toprule
Setpoints            & 0-4\,h & 5\,h-8\,h & 9\,h-12\,h & 13\,h-16\,h \\ \midrule
$C^*_{A}$ [kmol/m$^3$] & 1.5 & 8   & 4.3 & 7   \\ \bottomrule
\end{tabularx}
\end{table}

Both the LSQ and MLL setpoint-tracking objective formulations are solved via the automatic differentiation wrapper in \emph{autograd-minimize}~\citep{autogrd_minimize} and the \emph{SLSQP} algorithm of the python-based optimization library \emph{scipy-optimize} \citep{scipy2020pythonlibrary} in standard settings.

\begin{figure*}
    \centering
    \includegraphics[width=\textwidth]{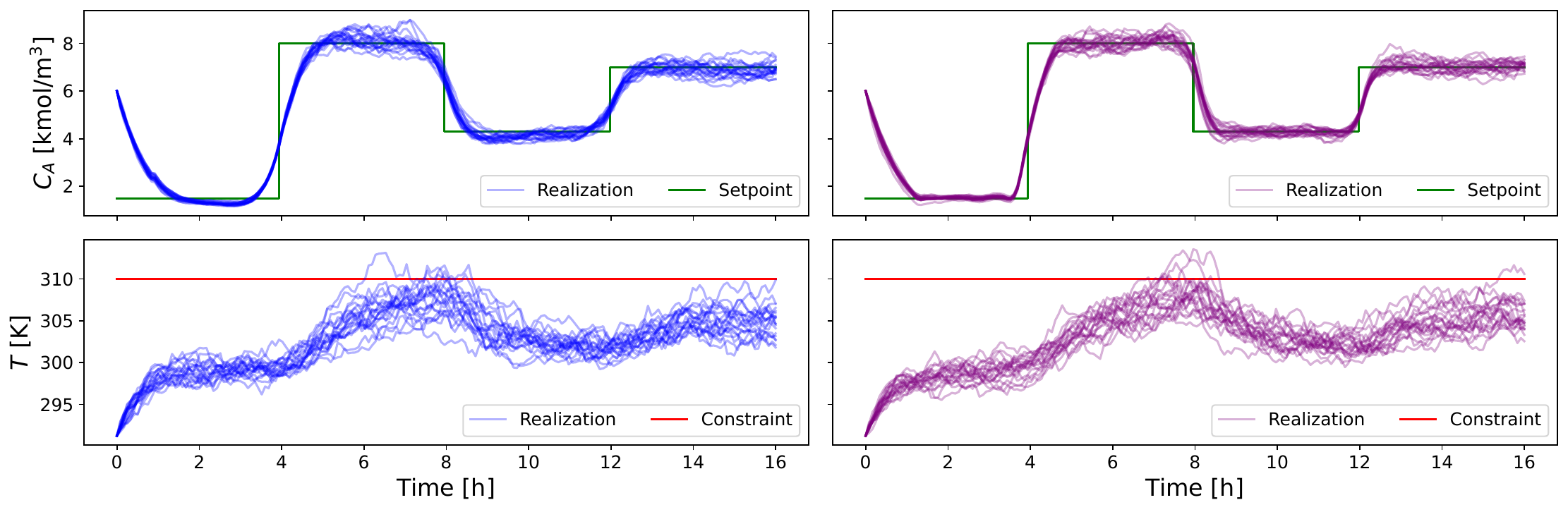}
    \caption{
    Realizations of reactor concentration $C_A$ (upper row) and reactor temperature $T$ (lower row) for open-loop optimization using the LSQ objective (left, blue) and the MLL objective (right, purple). 
    The setpoints are highlighted in green, and the constraint is shown in red. }
    \label{fig: open loop optimization scenarios}
\end{figure*}

Figure~\ref{fig: open loop optimization scenarios} shows 20 Euler-Murayama simulations for the results of the open-loop optimization using the LSQ objective (left, blue) and the MLL objective (right, purple), respectively. In addition, the setpoints for the reactor concentration $C^*_A$ are shown in green, and the constraint threshold on the reactor temperature is shown in red. 

The figure shows that both objective formulations lead to solutions that make the controller follow the setpoints. 
For both objectives, the reactor concentration quickly approaches the setpoints and then continues to fluctuate with the mean on the setpoints. 
The profiles for LSQ and MLL objectives show very similar behavior. 
However, it appears that the concentration profile resulting from the MLL controls takes slightly sharper control steps for the concentration $C_A$, see, e.g., at the end of the first control interval at 3.5\ h.

\begin{table}
\centering
\caption{Open-loop MPC error metrics and \%-age of maximum constraint violations. }
\label{tab: open loop MPC Errors}
\begin{tabularx}{\columnwidth}{@{}lXX@{}}
\toprule
           & LSQ   & MLL   \\ \midrule
MAE        & 0.49  & 0.44  \\
MSE        & 0.82  & 0.76  \\
Violations & 9.9\% & 7.7\% \\ \bottomrule
\end{tabularx}\end{table}
For a quantitative analysis, Table~\ref{tab: open loop MPC Errors} lists the mean-absolute-error (MAE), the mean-squared-error (MSE), and the \%-age of the maximum constraint violation for a simulation of 2000 scenarios of the solution shown in Figure~\ref{fig: open loop optimization scenarios}. 
The results for the error metrics confirm the observations of close matches of the setpoints in Figure~\ref{fig: open loop optimization scenarios}. Both MAE and MSE give low values for both objectives. 
Notably, the MLL objective results in slightly lower error metrics, which matches the observation of sharper control steps. 
The results for the maximum \%-age of constraint violations further confirm that the normalizing flow-based chance constraint formulation achieves good matches of the specified probability, i.e., the controls act neither too aggressively nor too conservatively.

In 2003, Kaut and Stein~\citep{kaut2003evaluation} published their seminal work on evaluating scenario generation methods for stochastic programs. 
They argue that scenario generation methods should yield bias-free and stable results in the optimization. 
Here, bias-free means that the solution of the scenario-based program should be identical to the solution for the true distribution of the random variable. 
They further define stability as a low variance in the solution of the scenario-based program, even for a small number of scenarios. 
Kaut and Stein refer to the objective function value in the optimum as the solution to the stochastic program. 

\begin{table}
\centering
\caption{Results of stability analysis for LSQ and MLL objectives~\citep{kaut2003evaluation}.
The table shows mean values ($\mu$), variance ($\sigma^2$), Kurtosis (KURT), and number of outliers (\# Outliers) for 100 stochastic programs for different scenario set sizes. 
}
\label{tab: Kaut Stein statistics}
\begin{tabularx}{\columnwidth}{@{}lXXXX@{}}
\toprule
                    & $S=3$  & $S=5$  & $S=10$ & $S=20$ \\ \midrule
$\mu_{LSQ}$         & 2.1   & 1.7  & 1.3  & 0.9  \\
$\mu_{MLL}$         & 59.4  & 58.2 & 58.1 & 57.6 \\
$\sigma^2_{LSQ}$    & 12.7 & 10.3 & 6.4  & 1.7  \\
$\sigma^2_{MLL}$    & 12.9 & 5.3  & 3.0  & 1.6  \\
$\text{KURT}_{LSQ}$ & 9.3  & 12.3 & 22.9 & 94.5 \\
$\text{KURT}_{MLL}$ & 3.4  & 2.6  & 3.6  & 5.1  \\
\# Outliers$_{LSQ}$ & 9    & 7    & 4    & 1    \\
\# Outliers$_{MLL}$ & 1    & 0    & 1    & 3    \\ \bottomrule
\end{tabularx}
\end{table}
Table~\ref{tab: Kaut Stein statistics} shows statistics of the solutions to the open-loop MPC problem for different numbers of scenarios for the MCMC approximation of the LSQ objective (Equation~\eqref{Eq: stochastic MPC objective}) and the MLL objective (Equation~\eqref{eq: MLL MPC}) computed for 100 distinct sets of scenarios each. 
The mean values of the objectives for LSQ and MLL show very different values. Note that this is expected as the two functions describe different properties. 
Neither objective can be evaluated for the true probability distribution, as it is unknown. Hence, an analysis of the bias is not possible as intended by Kaut and Stein.
For reference, the objective function values for a high number of 500 scenarios are 0.85 for the LSQ and 58.3 for the MLL objective. 
More important for the analysis of the stability of the solution are the higher-order statistics. 
The variance and kurtosis of the LSQ solutions are consistently higher compared to the MLL solution. These differences result from the higher number of outliers for the LSQ solution in all but the $S=20$ case. 
The $S=20$ case shows similar variance for both objectives, which indicates a very stable result for the MLL objective, considering the much higher mean value. 
The MLL objective does show more outliers for $S=20$. However, this is likely a result of a narrow distribution, as indicated by the low kurtosis. 
In summary, both objectives are able to achieve stable results. 
However, the LSQ requires more scenarios for stability, while the MLL objective yields satisfactory stability for as few as five scenarios.

In conclusion of this analysis, the normalizing flow-based open-loop MPC achieves good results for setpoint control. 
Throughout the analysis of error metrics, constraint violations, and solution stability, both objective function formulations yield good results even for relatively small scenario sets. 
Notably, the MLL objective slightly outperforms the established LSQ objective in the error and, notably, yields more stable results for small scenario sets.

\subsection{Closed-Loop MPC}\label{ssec: CSTR closed loop}
Finally, this section runs the normalizing flow-based MPC in a closed-loop and draws a comparison to MPC based on the nominal model (Equation~\eqref{Eqs: CSTR model}).
The closed-loop controllers are run with the setpoints listed in Table~\ref{tab: open-loop control setpoints} and a prediction horizon of 2\,h, i.e., 24 steps. 
All simulations are initialized with the same random seed to ensure comparability between the approaches. 
In addition to the stochastic behavior of the system in Equation~\eqref{eq: CSTR noise}, a measurement noise is added to the Euler Murayama simulation of the real behavior. The variance of the noise is 0.2 for the concentration and 2 for the temperature. 
The normalizing flow MPCs are solved using 20 scenarios. 

\begin{figure*}
    \centering
    \includegraphics[width=\textwidth]{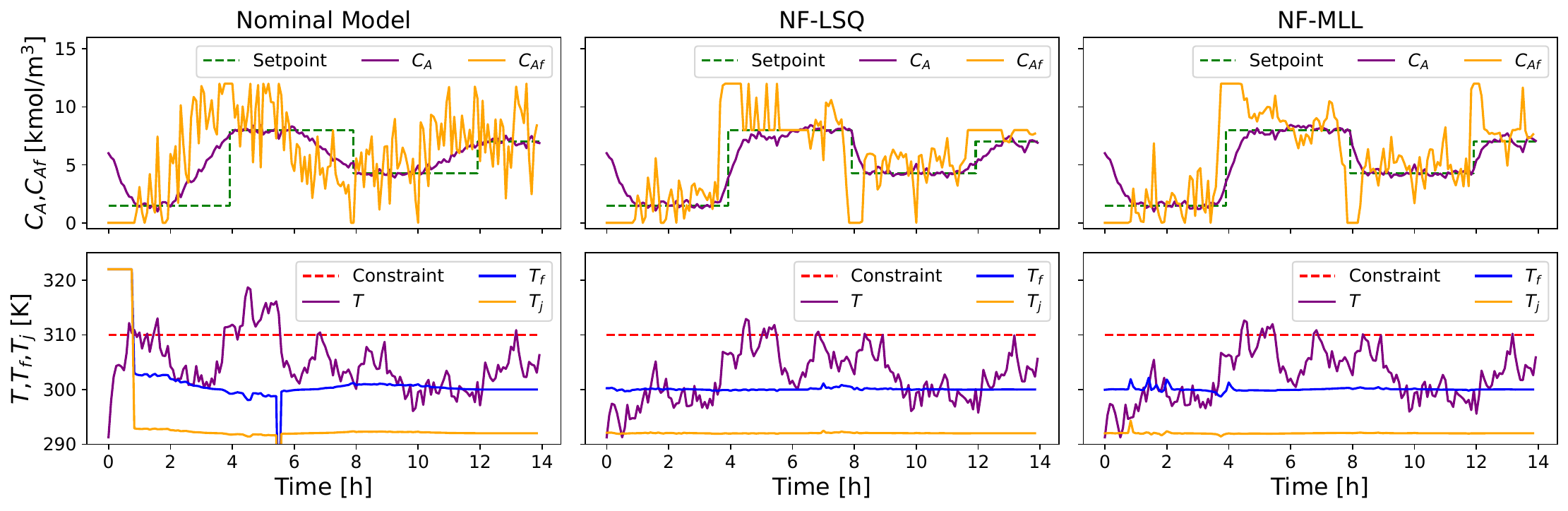}
    \caption{
    Closed-loop states and controls for nominal model (Equation~\eqref{Eqs: CSTR model}), the LSQ objective (NF-LSQ), and the MLL objective (NF-MLL). 
    The top column shows the concentration profiles, and the bottom column shows temperatures.
    The states are shown in purple, and the controls are shown in orange and blue. 
    }
    \label{fig: Closed-Loop MPC}
\end{figure*}
Figure~\ref{fig: Closed-Loop MPC} shows the CSTR states and control inputs for the nominal MPC (Nominal Model) and the normalizing flow MPC using the LSQ (NF-LSQ) and the MLL (ND-MLL) objective, respectively. 
Note that the results only cover 14 hours, as the prediction horizon for the MPC covers the last two. 
Similar to the open-loop case, the normalizing flow-based controllers tightly follow the concentration setpoint profile and quickly change between setpoints. 
The nominal model MPC also matches all setpoints, but shows slower transitions and higher errors.
Notably, the nominal MPC leads to higher and more constraint violations compared to the normalizing flow controllers. 
This result is expected as the nominal controller does not consider the stochastic behavior of the system. 
The neglect of randomness in the system is also evident in the larger and more frequent control increments by the nominal controller for the feed concentration $C_{Af}$. The MPC formulations in this work do not penalize the control steps and, thus, the nominal controller tends to take extreme action to react to the stochastic behavior. 
Meanwhile, the normalizing flow state space model considers the stochastic components of the system dynamics. Hence, both LSQ and MLL objectives lead to smoother control input profiles than the nominal model. 
Furthermore, the normalizing flow MPCs only show minor constraint violations, as already discussed for the open-loop controller. 
There are only minor differences between LSQ and MLL. The MLL results show slightly fewer control jumps in the feed concentration. Meanwhile, the state profiles are almost identical, where both achieve good setpoint tracking and low constraint violations.

\begin{table}
\centering
\caption{
Mean absolute error (MAE), mean squared error (MSE), and mean absolute control increment (MACI) for the closed-loop MPC. 
The MACS is computed only for the feed concentration $C_{Af}$. 
}
\label{tab: Errors closed-loop MPC}
\begin{tabularx}{\columnwidth}{@{}lXXX@{}}
\toprule
     & Nominal & LSQ  & MLL  \\ \midrule
MAE  & 1.1     & 0.55 & 0.49 \\
MSE  & 3.2     & 0.95 & 0.91 \\
MACI & 2.2     & 1.1  & 1.0  \\ \bottomrule
\end{tabularx}
\end{table}
Table~\ref{tab: Errors closed-loop MPC} shows the MAE, MSE, and the mean absolute control step (MACI) for the different closed-loop MPCs. 
The results for MAE and MSE confirm the observation from Figure~\ref{fig: Closed-Loop MPC} that the normalizing flow-based controllers track the setpoints better than the nominal controller.
Furthermore, the MACI for the nominal controller is double that of the normalizing flow-based controllers, again, confirming the observations from above.

In summary, the normalizing flow state space model leads to good results in the closed-loop MPC and outperforms the benchmark nominal controller in all evaluations, including state tracking under uncertainty and constraint violations. 
Again, the MLL objective shows slightly better results than the LSO objective. However, the results for LSQ and MLL are very similar and should not be interpreted to indicate a clear superiority of the MLL objective.

\section{Conclusion}\label{sec: conclusion}
This work applies the deep generative model called normalizing flows as a probabilistic state space model for chemical processes. 
The conditional PDF formulation of normalizing flows learns both the discrete-time stochastic dynamics with high flexibility, as there are no prior assumptions about the dynamics or their probability distribution. 

This work further compares LSQ and MLL formulations of the MPC objective. 
Both normalizing flow-based controllers outperform the nominal test case, and the results show similar results for both formulations, with slightly better results for the MLL objective. 
However, the results do not support a claim for clear superiority of the MLL at this point. 

Future work will investigate processes with more exotic stochastics, such as multimodal or other non-Gaussian distributions. 
Furthermore, analysis of the data efficiency will enhance confidence in the usability of the normalizing flow-based MPC for practical applications with limited data.

\appendix
\section{Energy Score}\label{app: energy score}
This work uses the energy score \citep{gneiting2008assessing, pinson2012evaluating} to assess the quality of the normalizing flow scenarios.
The energy score is defined as follows
\begin{equation}\label{Eq:Price_Delta_DefinitionEnergyScore}
    \begin{aligned}
    \text{ES}[k] = &
        \frac{1}{N_S} \sum_{s=1}^{N_S} \vert\vert \mathbf{x}[k] - \hat{\mathbf{x}}_s[k] \vert\vert_2\\
        &- \frac{1}{2{N_S}^2} \sum_{s=1}^{N_S} \sum_{s'=1}^{N_S} \vert\vert \hat{\mathbf{x}}_s[k] - \hat{\mathbf{x}}_{s'}[k] \vert\vert_2
    \end{aligned}
\end{equation}
Here, $\mathbf{x}$ is the realized state vector and $\hat{\mathbf{x}}_s$ are sample vectors generated using MCMC, $N_S$ is the number of samples, and $\vert\vert \cdot \vert\vert_2$ is the 2-norm.

\bibliographystyle{apalike}
  \renewcommand{\refname}{Bibliography}  \bibliography{References.bib}

\end{document}